\documentclass[nopreprintline,3p,times,11pt]{elsarticle}

\usepackage{setspace}
\usepackage{url}
\usepackage{multirow}
\usepackage[table,xcdraw]{xcolor}
\usepackage{tikz}
\usepackage{flushend}
\usepackage[export]{adjustbox}
\usepackage{comment}

\usepackage{tabularray}

\newtheorem{remark}{Remark}

\usepackage{amssymb}
\usepackage{amsmath}

\begin{document}

\begin{frontmatter}



\title{Over-the-Air Multi-Sensor Inference with Neural Networks Using Memristor-Based Analog Computing\tnoteref{label1}}

\tnotetext[label1]{Funding: This work is funded by The Scientific and Technological Research Council of Turkiye (TUBITAK), through European Coordinated Research on Long-term Challenges in Information and Communication Sciences and Technologies ERA-Net (CHIST-ERA) Project Sustainable Computing and Communication at the Edge (SONATA), under Grant CHIST-ERA-20-SICT-004 and TUBITAK Grant 221N366. \\
This article was presented in part at the IEEE Global Communications Conference (GLOBECOM), Kuala Lumpur, Malasia,
in December 2023 [DOI: 10.1109/GLOBECOM54140.2023.10436926].
}


\author[ins1]{Busra Tegin\corref{mycorrespondingauthor}}
\cortext[mycorrespondingauthor]{Corresponding author}
\ead{btegin@ee.bilkent.edu.tr}
\author[ins1]{Muhammad Atif Ali} 
\ead{atif@ee.bilkent.edu.tr}
\author[ins1]{Tolga M Duman} 
\ead{duman@ee.bilkent.edu.tr}

\affiliation[ins1]{organization={Department of Electrical and Electronics Engineering, Bilkent University},
            city={Ankara},
            postcode={06800}, 
            country={Turkey}}


\begin{abstract}
Deep neural networks provide reliable solutions for many classification and regression tasks; however, their application in real-time wireless systems with simple sensor networks is limited due to high energy consumption and significant bandwidth needs. This study proposes a multi-sensor wireless inference system with memristor-based analog computing. Given the sensors' limited computational capabilities, the features from the network's front end are transmitted to a central device where an $L_p$-norm inspired approximation of the maximum operation is employed to achieve transformation-invariant features, enabling efficient over-the-air transmission. We also introduce a trainable over-the-air sensor fusion method based on $L_p$-norm inspired combining function that customizes sensor fusion to match the network and sensor distribution characteristics, enhancing adaptability. To address the energy constraints of sensors, we utilize memristors, known for their energy-efficient in-memory computing, enabling analog-domain computations that reduce energy use and computational overhead in edge computing. This dual approach of memristors and $L_p$-norm inspired sensor fusion fosters energy-efficient computational and transmission paradigms and serves as a practical energy-efficient solution with minimal performance loss.
\end{abstract}


\end{frontmatter}



\section{Introduction} \label{WI_lit}

Deep neural networks (DNNs) are powerful but computationally complex machine learning models, which pose challenges for running DNN-based applications on simple edge units, such as Internet of Things (IoT) devices, due to their limited computational capabilities and energy constraints. To address this, pruning techniques can be used to remove insignificant weights and enhance performance and speed up learning and inference processes \cite{lecun1989optimal, hassibi1993optimal, blalock2020state}. 
{\color{black}{Pruning involves various techniques, such as layerwise pruning of weights with the lowest absolute values, pruning across the entire network, which enables faster inference by reducing network size. This process can be performed in two stages: train-time pruning, where pruning decisions are made simultaneously with each training update, and post-training pruning, where weights are pruned after model convergence. Train-time pruning, guided by a sparsity objective, is generally more efficient, although its implementation is more complex.}}
Another approach is to split the network between the edge device and a more powerful server, improving efficiency.

In \cite{shao_bottlenet_2020}, a network splitting approach is proposed where the authors introduce an end-to-end architecture called Bottlenet++ in which the encoder and decoder act as a machine learning-based joint source-channel coder, considering channel impairments as a parameter of the overall network. 
In \cite{Jankowski}, the authors study a similar scenario with feature compression and reliable communication via deep joint source-channel coding (DeepJSCC) over additive white Gaussian noise (AWGN) channels. 
In \cite{image_retrieval}, wireless image retrieval problem is considered where an edge device/sensor captures an image of an object as raw data, and transmits the corresponding low-dimensional signature with the aim of retrieving similar images belonging to the original object from another dataset. 
Wireless inference is also studied for graph neural networks (GNNs).
In \cite{Branchy-GNN}, the authors propose Branchy-GNN, a low-latency co-inference framework that uses network splitting and early exit mechanisms. In \cite{9606569}, the authors study decentralized inference with GNNs over imperfect wireless channels, with a focus on enhancing privacy in \cite{lee2022privacy}.
In references \cite{shao_task} and \cite{ shao_multi}, the information bottleneck (IB) principle \cite{tishby2000information, IB_deep} is used to formulate the trade-off between feature informativeness and inference performance for device-edge co-inference in task-oriented applications. 
{\color{black}{In \cite{9834591}, a distributed inference system is considered in which clients observe the same data and perform over-the-air ensemble inference using the decision vector, without network splitting, to ensure model privacy.}} 
As identified in earlier studies, wireless inference can be an effective solution to improve edge device inference capabilities while balancing communication overhead by utilizing computational resources at both the edge devices and the edge server \cite{li2018edge, liu2019edge,kang2017neurosurgeon,li2018jalad}. {\color{black}{The Metropolis-Hastings algorithm has also been applied to distributed averaging in wireless sensor networks \cite{8861554}.}}

DNN inference in traditional digital computing involves separate storage, memory, and compute elements \cite{von_neumann_first_1993}, which require continuous data transfer between them. This introduces bottlenecks, increasing both latency and energy consumption. In contrast, in-memory computing merges storage and computation on the same chip, eliminating the need for data shuffling between memory and processor. A critical operation in neural networks is the matrix-vector multiplication (MVM), where the inputs of a layer are multiplied by the weights of that layer to generate output preactivations. First characterized by Chua in 1971 \cite{chua_memristormissing_1971} and developed by HP Labs in 2008 \cite{strukov_missing_2008}, memristors are well-suited for MVM operations. Their conductance can be modulated by applying voltage pulses of varying amplitude, polarity, and duration. Memristors display several notable attributes, including nonvolatile memory, low latency, and reduced energy consumption \cite{zhang_brain-inspired_2020}. The use of memristor-based in-memory computing can enhance energy efficiency by up to a factor of $10^3$ compared to traditional CPU and GPU \cite{kosters_benchmarking_2023}. Explorations of memristors for applications such as storage \cite{isik_neural_2023} and in-memory computing \cite{sebastian_memory_2020} have demonstrated significant advantages for machine learning applications, particularly in energy-constrained edge devices.

In this paper, we examine sensor networks where multiple sensors gather data from overlapping regions and perform inference on a shared phenomenon. To reduce the computational burden of deep learning techniques, we split the network into two parts: a front-end on the sensor side and a back-end on a device with more computational power, {\color{black}{which can be considered a main server with virtually unlimited computational capacity and no energy or power restrictions, unlike the analog sensors that face both energy and computational limitations. Given the multiple sensors, this device also performs sensor fusion to obtain a single joint inference decision from various sensor observations.}}
{\color{black}{In multi-sensor networks, it is essential for the sensors to be low-cost and energy-efficient \cite{rault2014energy}, to minimize latency \cite{6182561}, to optimize data storage \cite{prauzek2018energy}, and to ensure privacy through straightforward encryption mechanisms \cite{khashan2021automated}.}} 
Hence, we explore the use of analog memristor neural networks for the sensors, addressing the challenges imposed by the computational demands and energy constraints prevalent in IoT edge devices. We demonstrate the effectiveness of the proposed approach by achieving acceptable performance using analog computations at the sensors, marking a significant advancement towards sustainable and efficient computation at the wireless edge.
Unlike previous works that focus on averaging operations for sensor fusion, our prior research \cite{tegin_transformation-invariant_2023} proposes using over-the-air (OTA) maximum approximations with LogSumExp and $L_p$-norm inspired sensor fusion techniques. This approach involves exact digital computations at the sensor side to enhance the usefulness of the gathered data \cite{teerapittayanon2017distributed, laptev2015transformation, laptev2016ti, mvcnn}. In this study, we explore a similar $L_p$-norm inspired function for feature fusion in memristor-based analog sensors. By leveraging multiple memristor-based analog sensors, we aim to achieve features that maximize overall inference accuracy in an energy- and bandwidth-efficient manner. Additionally, we introduce a trainable parameter to our $L_p$-norm approximation, allowing for customization of sensor fusion to better match the specific characteristics of the network.

Our main contributions are as follows:
\begin{itemize}
\item We introduce an $L_p$-norm inspired approximation to the maximum operation, which enables the extraction of transformation-invariant features. This approximation is performed in an over-the-air manner, making it bandwidth-efficient and helping to keep the transmission cost comparable to single-sensor setups.
\item We further introduce a learnable parameter, $p$, in our approach, which optimizes the sensor fusion method based on system specifics, such as the number of sensors, their placement, and data type. Additionally, due to the intermediate feature transmission from multiple sensors, our proposed setup offers greater privacy and security compared to raw data transmission.
\item We incorporate in-memory computing for sensors, using analog computations at the sensor side for the front-end computations of the network. This enhances energy efficiency compared to traditional CPU and GPU computations, with only minimal performance sacrifice, offering an efficient solution for complex tasks in distributed environments.
\item We validate our proposed approaches with extensive simulations using the custom-made Car Learning to Act (CARLA), ModelNet, and MNIST datasets under various setups and channel conditions.
\end{itemize}

The paper is organized as follows. Section \ref{lernable_sm} introduces the system model. Over-the-air sensor fusion for multi-sensor wireless inference is introduced in Section \ref{sec:otaFusion}, while in-memory computing and its application to multi-sensor inference are presented in Section \ref{sec:in-memory}. We summarize the inference and training processes of multi-sensor wireless inference systems with memristors in Section~\ref{sec:training}, while the corresponding calculations for inference energy consumption are provided in Section~\ref{sec:energy_consumption}. Performance of several tasks over wireless channels are studied via simulations in Section \ref{sec:numEx}, and the paper is concluded in Section \ref{sec:conc}.

\section{System Model} \label{lernable_sm}

\begin{figure}
    \centering
    \includegraphics[width=0.75\linewidth]{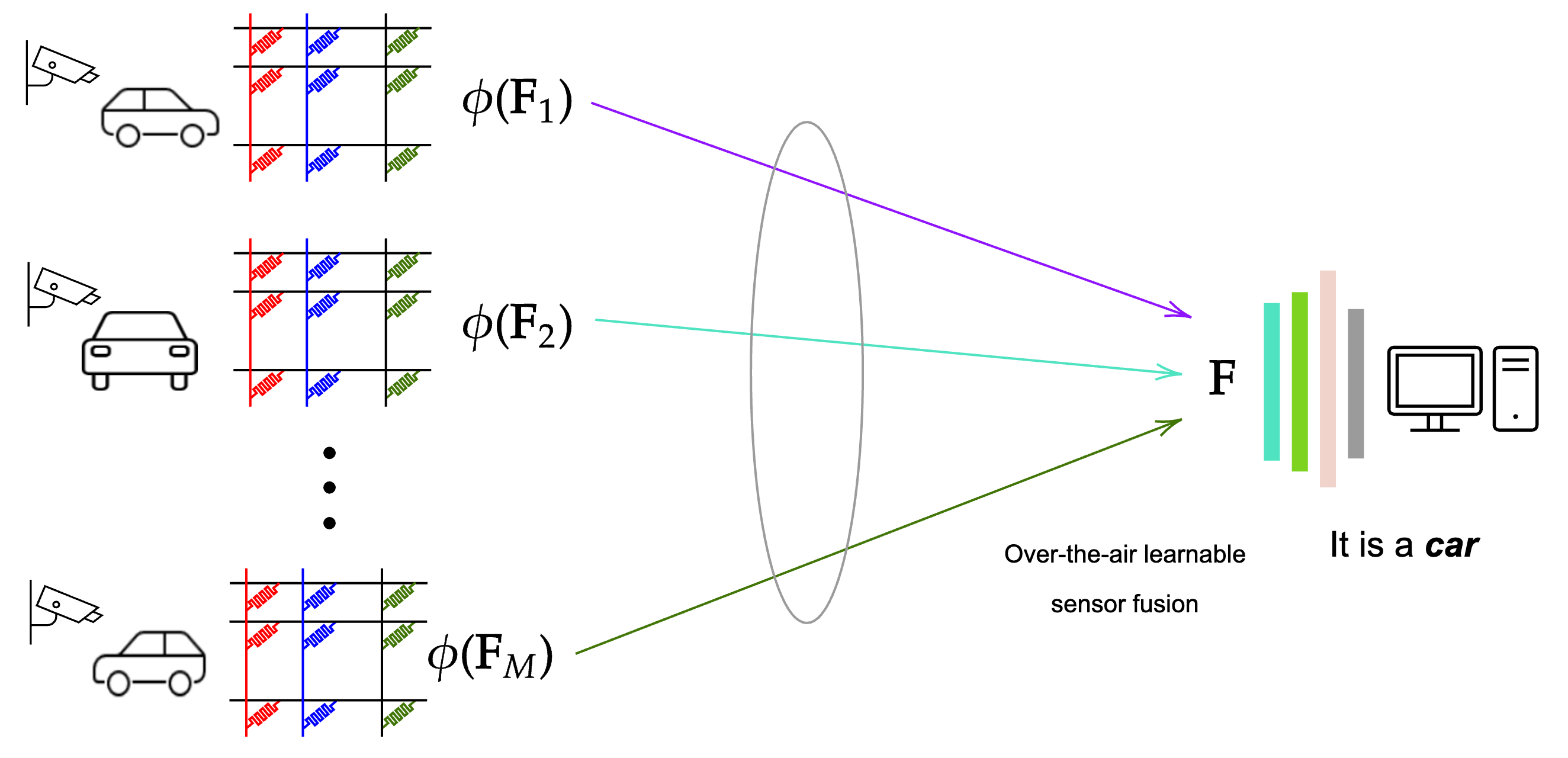}
    \caption{System model for the multi-sensor wireless inference with analog memristor neural networks.}
    \label{fig:system_model}
\end{figure}

We consider a wireless sensor network with $M$ analog sensors that observe the same phenomena for a common object. With the help of the multi-sensor structure, an inherent data augmentation is introduced to increase the inference reliability. Specifically, we develop both transformation-invariant and learnable OTA sensor fusion techniques, as depicted in Fig. \ref{fig:system_model}. Each analog sensor $m \in \{1, \cdots, M\}$ processes its own detected raw data using a front-end network that uses in-memory computation made possible by a memristive crossbar chip, to obtain the intermediate feature vector, $\mathbf{F}_m$. 
{\color{black}{Note that $\mathbf{F}_m$ can be 1D vectors, 2D matrices or 3D tensors depending on the network structure and layers. For simplicity, we assume $\mathbf{F}_m \in \mathbb{R}^d$ is a one-dimensional vector that can be obtained by flattening 2D matrices or 3D tensors where $d$ depends on the input sample, and network parameters.}} 
The intermediate feature vector is then preprocessed by the function $\phi(\cdot)$ and transmitted over a wireless channel to a central device to complete the inference. 
{\color{black}{The preprocessing function, $\phi(\cdot)$, for each sensor $m$ takes the intermediate feature vector $\mathbf{F}_m$ as input and outputs another vector without changing its size. A simple example of $\phi(\cdot)$ is to directly output the input vector, resulting in averaging as a form of sensor fusion when over-the-air transmission is used. Another straightforward example is to apply a power operation to $\mathbf{F}_m$, which either amplifies or diminishes specific elements of the feature vector depending on the power. By selecting specific power values or optimizing it, this parameter can also enable different sensor fusion techniques, such as averaging or element-wise maximum, as we will describe in the following sections.}} This approach allows an implicit collaboration between the sensors and central device for collaborative inference in multi-sensor networks.

With the proposed approach, the activations of each sensor are transmitted over an AWGN multiple access channel (MAC) and combined over-the-air, leading to an efficient solution that decreases transmission overhead and bandwidth requirements while using a single helper device. 
{\color{black}{AWGN is a basic random noise model used for communication channels. It represents the linear addition of white noise with constant spectral density and Gaussian distribution to the source signal. This model can effectively represent background noise in a communication channel, arising from various sources such as thermal noise in electronic components, atmospheric noise, or interference from other devices, all of which can degrade channel capacity due to added noise. To mathematically model the channel between the sensors and the device, we incorporate this noise model. Additionally, to reduce transmission costs, we utilize a MAC channel enabling over-the-air transmission.}} 
The goal of the over-the-air combining operation is to obtain a single, reliable and representative feature vector from multiple sensors, instead of a separate vector for each sensor. This approach only requires the transmission of feature vectors on a shared link and the central device aims to recover the combination of these features.  Therefore, our system requires less bandwidth compared to traditional methods, which require separate transmission and recovery of each signal.

To simulate memristive behavior for computation in an analog neural network, multiple models are presented in the literature \cite{carbajal_memristor_2015}. We base our results on a phase change memory (PCM) model developed in \cite{nandakumar_phase-change_2018} and \cite{noauthor_inference_nodate} due to its ease of implementation and foundation on empirical data from real measurements.

\section{Over-the-Air Sensor Fusion for Multi-Sensor Wireless Inference}\label{sec:otaFusion}

In wireless networks, over-the-air aggregation is a technique used to improve spectral efficiency by allowing multiple transmitters to share the same communication resources. In this technique, the transmitters send their data to the receiver simultaneously, and the receiver can directly obtain the summation or average of the transmitted symbols without explicitly decoding the individual symbols using the superposition property of the MAC. This property has been exploited in various wireless communication scenarios, including over-the-air computation scenarios (see, e.g., \cite{9535447, 9382114, 10018930, 8371243}), which can also be used as a sensor fusion method for multi-sensor inference setups. However, there is no guarantee that using feature averaging will result in the best inference accuracy.

Acquiring transformation-invariant features is indispensable for ensuring reliable outcomes across diverse data types, including image, video, sound or radar data, when faced with alterations such as rotations, crops, time shifts, amplitude changes, phase shifts or other variations that are inherent to each domain.
A simple yet effective solution is data augmentation, as presented in \cite{van2001art}. A more sophisticated approach is a transformation-invariant (TI) pooling operator \cite{laptev2015transformation}, implemented in \cite{laptev2016ti}. For instance, for vision problems, this method feeds different rotated versions of the same sample into the first part of the network, and combines them using a transformation invariant (TI)-pooling layer to perform the remaining operations required for the network. This TI-pooling layer implements an element-wise maximum operator, resulting in a transformation-invariant feature vector. In \cite{mvcnn}, the authors design a multi-view convolutional neural network (MVCNN) for 3D shape recognition using 2D section images of the 3D model, using the maximum operator to pool and combine layers in a similar manner.  
Standard maximum operation requires the receiver to recover each transmitted signal separately before taking the element-wise maximum. 
However, this process demands a significant amount of bandwidth and can also be computationally expensive, making it infeasible for real-time applications.
In this paper, we are interested in transformation-invariant over-the-air sensor fusion for multi-sensor networks; however, to improve the representativeness of the fused feature vector, we aim to move beyond simple averaging techniques. 
To enable an approximate maximum operation for the multi-sensor wireless inference setup in a bandwidth efficient manner, we propose using an $L_p$-norm inspired function. These approximation allows us to approximate the maximum operation using a combination of elementary functions that are computationally efficient to implement. By leveraging the superposition property of the MAC, we can exploit the benefits of maximum combining without sacrificing spectral efficiency while improving the overall inference performance and robustness of the system. 

Despite these advantages, practical implementation of transformation-invariant features encounters challenges, such as the need to position sensors throughout the entire scene and to identify a set of transformations that satisfy the necessary properties. These limitations impose constraints on wireless inference systems, and even with transformation-invariant features, achieving the highest accuracy is not guaranteed. To address these issues, we further propose to use a trainable sensor fusion function to optimize feature fusion and enhance the overall inference accuracy. This function incorporates a trainable parameter that provides flexibility in sensor fusion based on system properties. By adjusting this parameter, a range of behaviors can be captured, from feature averaging to approximating the maximum operation.

It is worth noting that while the maximum operation is utilized in \cite{laptev2016ti} to train a robust network for rotation or scale changes, and \cite{mvcnn} focuses on computationally efficient 3D shape recognition using 2D views, our study takes a unique approach by considering a distributed multi-sensor network rather than relying on a single machine for computations and data sensing. Specifically, we explore the use of locally obtained data from each sensor, which is then combined in a bandwidth-efficient way with the approximate maximum operation for further processing and joint inference, distinguishing our work from the aforementioned studies.

\begin{remark}
Note that, unlike distributed and federated learning, this study focuses on the inference phase. Our approach involves offline training, which can be performed on a powerful device beforehand, followed by sharing the network branch weights between the sensors. The sensors use these pre-trained network weights for analog computations, while the central device handles real-time inference. It is worth noting that although the pre-trained network weights are shared with the sensors before the inference phase, there is no explicit communication among the sensors during inference. This makes the proposed approach highly scalable and suitable for large-scale networks with many analog sensors while reducing energy consumption with the help of memristors. Overall, our approach provides an effective solution for achieving real-time, bandwidth- and energy-efficient inference using multiple memristor-based sensors in a network.
\end{remark}

\subsection{$L_p$-norm inspired approximation for over-the-air maximum} \label{ssec_Lp}
Since we are interested in obtaining the maximum of the transmitted features to get the transformation-invariant features in a multi-sensor network with OTA sensor fusion, the $m$-th worker deploys the front-end of the network resulting in  feature vector $\mathbf{F}_m$. Instead of directly transmitting $\mathbf{F}_m$, each sensor will deploy the preprocessing function $\phi(\cdot)$, then the $m$-th sensor will obtain and transmit 
\begin{equation} \label{Lpx}
    \mathbf{x}_m = \phi(\mathbf{F}_m) = \mathbf{F}_m^p,
\end{equation}
for the $L_p$-norm inspired approximation where $p > 0$\footnote{The function given in \eqref{Lpx} is inspired by the $L_p$ norm but is adapted for usage when $p > 0$, diverging from the conventional $L_p$ norm definition for $ 0 < p < 1$.}. 
It is worth mentioning that $\mathbf{F}_m^p$ can be expressed as a vector consisting of $d$ elements, denoted by $[\mathbf{F}_m(1)^p, \mathbf{F}_m(2)^p, \cdots, \mathbf{F}_m(d)^p]$, where $\mathbf{F}_m(i)$ is a scalar representing the $i$-th element of vector $\mathbf{F}_m$.
{\color{black}{For instance, if we use a fully connected neural network and cut the front-end at layer $l$ with $d$ neurons, the output of this layer will be the intermediate feature vector of scalars with a length of $d$ and $\mathbf{F}_m(i)$ is the $i$-th element of this vector.}} 
Note that $L_{\infty}$ provides the largest magnitude among each vector element; hence having larger $p$ enables a better approximation of the maximum. 
We further emphasize that $\mathbf{F}_m$'s are the outputs of commonly used activation functions (e.g., rectified linear unit (ReLU) and sigmoid). Hence, the approximation for the largest magnitude will be an approximation for the maximum. 

The received signal at the device is 
\begin{align} \label{Lp1}
    \mathbf{y} = \sum_{m=1}^M \mathbf{x}_m + \mathbf{n} = \sum_{m=1}^M  \mathbf{F}_m^p + \mathbf{n},
\end{align}
where $\mathbf{y} \in \mathbb{R}^d$, and $\mathbf{n} \in \mathbb{R}^d $ is the AWGN noise vector with variance $\sigma_n^2$. 

The $i$-th element of the received signal \eqref{Lp1} is
\begin{align} \label{LSE2}
    {y}(i) = \sum_{m=1}^M  {F}_m(i)^p + n(i),
\end{align}
for $i \in \{1, \cdots, d\}$. Before processing the received signal at the back-end part of the network, one needs to take the $\frac{1}{p}$-th power of the received signal which is 
\begin{align} \label{eq_post}
    \max\{F_1(i), \cdots, F_M(i)\} &\approx 
   {y}(i)^{\frac{1}{p}} \nonumber \\
    &= \left(\sum_{m=1}^M  {F}_m(i)^p + n(i)\right)^{\frac{1}{p}}.
\end{align}
This operation provides an approximation for the element-wise maximum and the resulting vector is fed to the back end of the network which will complete the inference operation based on the fused feature vector. 

{\color{black}{To better understand the sensor fusion function \( \phi(\mathbf{F}_m) \) used in \eqref{eq_post}, let us consider two extreme ends of the spectrum. When \( p = 1 \), this is equivalent to performing over-the-air transmission without additional processing, resulting in the summation (or equivalently, the same information of average operation) of the transmitted signals at the receiver. However, as \( p \) increases, applying the preprocessing function to nonnegative feature vectors causes the feature value of the sensor with the largest magnitude to dominate the others. Thus, this approach approximates the maximum operation using the described preprocessing and postprocessing functions. We further note that for large but finite \( p \), this method provides a close approximation of the maximum.
}}

\subsection{Learnable $L_p$-norm inspired over-the-air sensor fusion}

In \cite{laptev2015transformation, laptev2016ti}, it is suggested that transformation-invariant features can be derived from multiple data sources through a maximum operation, provided that all potential input transformations form a group adhering to fundamental properties such as closure, associativity, invertibility, and identity. However, practically situating sensors throughout the entire scene to meet these conditions may be infeasible. Furthermore, identifying a set of transformations that fulfills all the required properties can be a daunting task, especially given the potentially vast number of possible transformations. These limitations impose practical constraints on wireless inference setups. Moreover, it is essential to note that even with transformation-invariant features, achieving the highest possible inference accuracy is not guaranteed.

To overcome these challenges, we adopt a similar approach to the previous section, using the same preprocessing and postprocessing functions \eqref{Lpx} and \eqref{eq_post}, where a constant $p \geq 0$ is used to approximate the maximum operation as a sensor fusion technique.
Here, we further introduce the use of a trainable parameter $p$ instead of a constant. This approach allows us to capture a spectrum of behaviors by providing flexibility for sensor fusion based on system properties. Specifically, when $p = 1$, the function corresponds to feature summation, conveying the same information as averaging, while for sufficiently large values of $p$, it approximates the maximum operation. During training, this parameter can converge to an optimal value that maximizes inference accuracy for the specific setup.

During offline training, the parameter $p$ of the sensor fusion method is introduced as a trainable network parameter. Consequently, this parameter, in conjunction with the network parameters, undergoes adjustments to minimize a suitable loss function, thereby fine-tuning the entire network. This optimization process depend on various aspects, including the number of sensors, their quality, placement, and the sensor fusion method, all adjusted for the specific characteristics of the sensor network.
After the offline training process, the resulting value of parameter $p$ is fixed, and this fixed value is used during the real-time online inference process.

{\color{black}{The multi-sensor wireless inference system with sensor fusion using the $L_p$ norm with trainable $p$ not only reduces transmission costs through over-the-air transmission but also enables a joint inference result from shared phenomena. In traditional transmission schemes, transmitters occupy separate time slots, requiring sensors to be assigned to their respective slots. Joint inference can only be performed after all transmitters have completed their transmissions, resulting in transmission latency that is proportional to the number of sensors. In contrast, over-the-air transmission allows all transmitters to use a single shared time slot and transmit simultaneously, significantly reducing transmission time. However, performing this inference using digital computing has several drawbacks: it requires separate storage, memory, and compute elements, which results in continuous data transfer between them, leading to increased latency and energy consumption \cite{von_neumann_first_1993}.

As an alternative to digital computing, we introduce the use of in-memory computing within the multi-sensor wireless inference setup. This approach integrates storage and processing on the same chip, thereby eliminating the data shuffling between memory and the processor. Memristors, known for their nonvolatile memory, low latency, and energy-efficient operation, are highly suitable candidates for implementing in-memory computing for multi-sensor wireless inference systems. Despite their use in various studies related to machine learning and the presentation of different models for their mathematical representation \cite{carbajal_memristor_2015, nandakumar_phase-change_2018, noauthor_inference_nodate}, to the best of our knowledge, this is the first application of memristors in a multi-sensor wireless inference setup. Combined with our novel over-the-air learnable sensor fusion strategy, the employment of memristors significantly reduces energy consumption and latency, making the proposed system more feasible for real-world scenarios.

}}

\section{In-Memory Computing and Its Application to Multi-Sensor Inference} \label{sec:in-memory}
The approach described in the previous section ensures the transmission efficiency of a multi-sensor network by utilizing a transformation-invariant and trainable OTA sensor fusion approaches. In addition to this improvement, the energy efficiency of the overall system can be further enhanced by employing memristors for neural network operations. Note that in-memory computing using memristors can be applied to both the front-end and back-end of a neural network. However, in this study, we focus solely on the front-end network; specifically, we assume that analog sensors with memristive neural networks are employed, while the proposed approach can be easily generalized to fully analog networks. It is important to note that the use of memristors may introduce additional impairments during the process of obtaining intermediate features due to imperfect and noisy operations. To accurately model memristive computations, we adopt the computation models described in \cite{nandakumar_phase-change_2018} and \cite{noauthor_inference_nodate} for PCM-based memristors. This model identifies three types of noise elements that occur during weight updates and readings, as well as a time-dependent decay of programmed values.

Programming noise: This noise originates from discrepancies between the target and the actual conductance values programmed in the hardware. This is represented as an additive normal distribution with zero mean and standard deviation $\sigma_\text{prog}$.  The conductance after programming is then given as
\begin{equation}\label{eqn:n_prog}
    g_\text{prog} = g_T + \mathcal{N}\big(0, \sigma_\text{prog}\left(g_T\right)\big),
\end{equation}
where $\sigma_\text{prog}\left(g_T\right)$ is defined as a quadratic function of target conductance $g_T$, which is
\begin{align}\label{eqn:sigma_prog}
    \sigma_\text{prog}(g_T) = \max( -1.1731g_T^2+1.9650g_T  + 0.2635, 0).
\end{align}

Drift noise: Conductance drift is an intrinsic property of the phase-change material and is due to structural relaxation of the amorphous phase \cite{le_gallo_collective_2018}. Knowing the conductance $g_\text{prog}$ programmed at time $t_c$, the conductance evolution $g_\text{drift}\left(t\right)$ can be modeled as a decaying exponential with exponent $\nu$ as
\begin{equation}\label{eqn:g_drift}
    g_{\text{drift}}\left(t\right) = g_\text{prog}\left(\frac{t}{t_c}\right)^{-\nu},
\end{equation}
where $\nu$ follows a normal distribution whose mean $\mu_\nu$ and standard deviation $\sigma_\nu$ which are the functions of target conductance $g_T$, and defined as
\begin{align}
\nu &= \mathcal{N}\big(\mu_{\nu}\left(g_T\right),\sigma_{\nu}\left(g_T\right)\big), \\
    \mu_{\nu}(g_T) &= \min\big(\max \left(-0.0155 \log{g_T} + 0.0244, 0.049\right),  0.1\big), \\
    \sigma_{\nu}(g_T) &= \min\big(\max \left(-0.0125 \log{g_T} - 0.0059, 0.008\right),   0.045\big).
\end{align}

Read noise: This noise source appears during the execution of matrix vector multiplications with the in-memory computing hardware. It stands for instantaneous fluctuations in hardware conductance due to the intrinsic noise characteristics of PCM devices, namely $1/f$ noise and random telegraph noise \cite{fugazza_random_2010}. This is an additive Gaussian noise whose standard deviation is a function of the drift conductance defined in \eqref{eqn:g_drift}, a coefficient $Q_s$, and the current read time $t$. This noise then leads to further fluctuation in the total conductance value.
The coefficient $Q_s$, standard deviation  $\sigma_{nG} (t)$ of the read noise and the final conductance value $g(t)$ are given as
\begin{align}
    Q_s =& \min\left(\frac{0.0088}{g_T^{0.65}}, 0.2\right), \\
    \sigma_{nG} (t) =&  g_{\text{drift}}\left(t\right) Q_s \sqrt{\log{\frac{t + t_c}{2t_{\text{read}}}}}, \\
    g\left(t\right) = & g_{\text{drift}}\left(t\right) + \mathcal{N}\left(0, \sigma_{nG} \left(t\right)\right).
\end{align}

Conductance Mapping: To translate the trained weight values into conductance terms, a differential configuration employing a pair of memristors is used. Each weight, represented as $w_{ij}$, is programmed onto two PCM devices. Depending on the sign of the weight, one device is programmed while the other is set to its minimum conductance state $g_{\min}$. The target conductance for the positive and negative memristors $g_{T_\text{pos}}$ and $g_{T_\text{neg}}$, respectively, is then determined by
\begin{align}
    g_{T_\text{pos}} &= 
    \begin{cases}
        w \frac{g_{\max} - g_{\min}}{\max\left(\left|\mathbf{W}\right|\right)} & \text{if $w>0$} \\
        0 & \text{otherwise},
    \end{cases}
    \\
    g_{T_\text{neg}} &= 
    \begin{cases}
        -w \frac{g_{\max} - g_{\min}}{\max\left(\left|\mathbf{W}\right|\right)} & \text{if $w<0$} \\
        0 & \text{otherwise}.
    \end{cases}
\end{align}
Next, the following noise models are applied:
\begin{align}
  g_{T_{pos}}' &= \left( g_{T_{pos}} + \mathcal{N}\left(0, \sigma_\text{prog}\left(g_{T_{pos}}\right)\right)\right)\left(\frac{t}{t_c}\right)^{-\nu}  + \mathcal{N}\left(0, \sigma_{nG} \left(t\right)\right), \\
  g_{T_{neg}}' &= \left( g_{T_{neg}} + \mathcal{N}\left(0, \sigma_\text{prog}\left(g_{T_{neg}}\right)\right) \right)\left(\frac{t}{t_c}\right)^{-\nu}  + \mathcal{N}\left(0, \sigma_{nG} \left(t\right)\right).
\end{align}
These \textbf{noisy} conductance values are then reverse-mapped to obtain the noisy weights as
\begin{gather}\label{eqn:w_noisy}
  w' = \frac{\left( g_{\max} - g_{\min} \right) \left( g_{T_{pos}}' - g_{T_{neg}}' \right)}{{\max\left(\left|\mathbf{W}\right|\right)}}.
\end{gather}
Here, $w'$ symbolizes the noisy weights after factoring in the various noise sources of the PCM memristive devices.

{\color{black}{
\section{Multi-Sensor Inference with In-Memory Computing and Training Process} \label{sec:training}
}}


As discussed previously, $L_p$-norm inspired sensor fusion enhances transmission efficiency via OTA transmission, while in-memory computing significantly reduces energy consumption. Hence, combining these approaches can create a communication-efficient system suitable for energy-constrained edge devices.


In our proposed solution, the inference phase is performed in real-time using a pre-trained neural network, which employs the forward pass on analog memristive sensors. We simulate in-memory inference using PCM-type memristors by replicating and mapping the shared sensor weights to all $M$ sensors as described in Section \ref{sec:in-memory}. We add the PCM noise during the forward pass and compute a noisy activation for the $m$-th sensor as
\begin{equation}\label{eq:noisy_activation}
    \textbf{F}_m = f\left(w', \mathbf{u}_m\right),
\end{equation}
where $w'$ are the network weights after the addition of noise as described in \eqref{eqn:w_noisy} and $\mathbf{u}_m$ is the input for the $m$-th sensor. This setup simulates a real-world scenario where each sensor experiences a unique weight distribution after programming. For a single-sensor scenario, we start the training with random initialization. As the number of sensors increases, we use the model trained for the previous case, e.g., the training with $m$ sensors is initialized with the network trained for $m-1$ sensors.

{\color{black}{
To emulate real-world conditions, we simulate the behavior of PCM-type memristors during inference. The shared sensor weights are replicated and mapped to all $M$ sensors. PCM noise, an inherent characteristic of memristive devices, is modeled during inference by perturbing the network weights, simulating scenarios where each sensor experiences unique weight variations post-programming.

Additionally, to handle multiple sensors effectively, we adopt a sequential training paradigm. Starting with a single sensor, the network is trained with randomly initialized weights. For configurations with additional sensors, the training for $m$ sensors is initialized using the model trained for $m-1$ sensors. This approach allows the network to adapt incrementally to the increased complexity of multi-sensor scenarios.

In our preliminary experiments, we investigated noise-aware training methods \cite{kariyappa2021noise} by incorporating PCM noise during the training phase to improve robustness against hardware variability. However, the initial results showed minimal improvements in inference accuracy for this specific multi-sensor setup. This outcome suggests that in scenarios where memristor noise is not the dominant factor—such as cases where channel noise or other external sources are more significant—noise-aware training requires further investigation to fully understand its impact and effectiveness.
}}

\begin{remark}
   We note that due to the computational limitations of edge devices, the training is performed offline without PCM noise. The weights and parameter $p$ of sensor fusion are shared with the sensors and the central device for real-time inference.
\end{remark}

{\color{black}{
\section{Memristor-Based Energy Consumption}\label{sec:energy_consumption}

The energy efficiency of sensors is a critical factor in the design of distributed inference systems, particularly for IoT and edge computing applications where resources are constrained. In this section, we provide a framework for analyzing the energy consumption of memristor-based sensors and compare it with conventional digital systems for the multi-sensor wireless inference.  This analysis builds on the methodology introduced for single-sensor memristor networks in our primary work \cite{ali2024learning}.

Memristor-based sensors leverage in-memory computing to perform operations such as MVM directly within the memory elements, eliminating the energy costs of data transfer between memory and processing units. The energy dissipated during inference depends on the programmed conductance of the memristor crossbar, the input voltage, and the duration of the operation. 

For the $m$-the memristor-based sensor, the energy consumed during an MVM operation can be expressed as:
\begin{equation}\label{eq:energy_sensor_single}
    E_m = {\lVert \mathbf{G}_m \mathbf{V}_m^2 \rVert}_1 t,
\end{equation}
where $\mathbf{G}_m$ is the conductance matrix representing the memristor crossbar, $\mathbf{V}_m$ indicates element-wise squaring of the input voltage vector, $t$ is the duration of the operation, and ${\lVert \cdot \rVert}_1$ denotes the sum of the absolute values of the elements in the resulting vector.

For a system with $M$ sensors operating in parallel, the total energy consumed by the sensors during one inference operation is:
\begin{equation}\label{eq:energy_sensors_total}
    E_\text{total} = \sum_{m=1}^M E_m,
\end{equation}
where $E_m$ is given in \eqref{eq:energy_sensor_single}.

\subsection{Upper Bound on Energy Consumption}
Calculating the exact energy consumption is challenging; therefore, we opted to estimate an upper bound. To determine the maximum energy consumption for a memristor-based system, we assume that all memristor conductances are set to their maximum value, $g_\text{max}$, and all input voltages are at their maximum value, $v_\text{max}$.

Under these conditions, the upper bound on the total energy consumed across all sensors is given by:
\begin{equation}\label{eq:upper_bound_sensors}
    E_\text{total}^{\text{max}} = M \times (\text{number of crossbar elements per sensor}) \times g_\text{max} \times v_\text{max}^2 \times t.
\end{equation}
This upper bound provides a worst-case estimate of energy consumption, offering insights into the scalability and feasibility of memristor-based systems in large-scale deployments.


\subsection{Comparison with Digital Sensors}

Digital sensors, such as Raspberry Pi (RPi)-based platforms, are widely used in edge inference setups due to their flexibility and general-purpose capabilities. However, digital systems typically consume constant power regardless of workload because they run an operating system (OS) and multiple background processes. 

The energy consumed by a digital sensor during an inference operation lasting $t$ seconds can be expressed as:
\begin{equation}\label{eq:energy_rpi}
    E_\text{RPi} = P_\text{RPi} \times t,
\end{equation}
where $P_\text{RPi}$ is the power consumption of the Raspberry Pi, typically around 15 W during inference. Unlike memristor-based systems, which consume energy only during active operations, digital sensors exhibit constant power consumption irrespective of input activity. This fixed power usage can lead to significantly higher energy costs in scenarios with sparse or intermittent input data.


}}

\section{Numerical Results}\label{sec:numEx}
In this section, we present numerical examples using various distributed inference scenarios with over-the-air sensor fusion, employing both analog and digital computations across multiple datasets.

\subsection{Dataset descriptions}
In our numerical examples, we utilize three datasets: 1) the ModelNet dataset, 2) a custom-made CARLA dataset, and 3) rotated MNIST dataset
for the performance evaluation of the proposed approaches.

\begin{figure}[b]
\centering
\includegraphics[width=11cm]{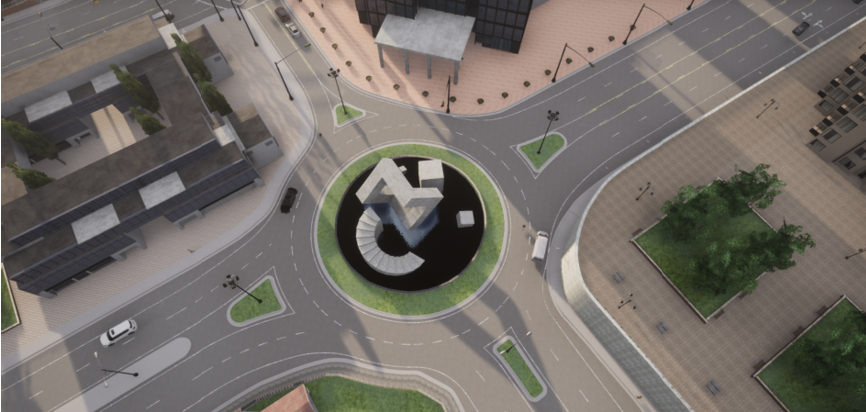}
\caption{Roundabout simulation environment in CARLA.}\label{CARLA_sm}
\end{figure}

\textbf{ModelNet}: The Princeton ModelNet dataset, specifically its 40-class subset \cite{Wu_2015_CVPR}, originally provides 3D CAD models of objects. These models have been processed using the method proposed in \cite{mvcnn} to obtain 2D samples from a 12-view camera setup. In our simulations, we utilize this 2D dataset to assess the performance of our proposed approach.

\textbf{CARLA}: Wireless inference systems have gained traction in various domains, from autonomous driving to surveillance. These systems rely on sensor data fusion for accurate environmental interpretation. However, evaluating these systems in real-world scenarios is challenging due to complexity and cost. To address this, we use the CARLA platform \cite{dosovitskiy2017carla} to create a simulation environment.
Our setup includes a roundabout scenario with eight strategically placed cameras (as in Fig. \ref{CARLA_sm}). We use data from these cameras to construct a robust dataset that simulates wireless communication challenges like occlusions, signal variations, and interference. This dataset is crucial for training and evaluating wireless inference algorithms, enabling accuracy, robustness, and efficiency analysis in a simulated, yet highly realistic, environment. It also allows for algorithm comparisons, suitability assessments, and the development of new approaches.

In this environment, we collect training and test data for five classes: pedestrians, small vehicles, large vehicles, bicycles, and motorcycles. Table \ref{tab_carla} summarizes the dataset properties.
It is important to note that although there are eight cameras positioned around the roundabout, some detected objects may only be sensed by a subset of the cameras. Consequently, the number of collected images for a particular object class could be less than eight.

\begin{table}[]
\centering
\caption{The dataset properties for the custom-made CARLA dataset.}\label{tab_carla}
\vspace{0.15cm}
\begin{tabular}{|l|c|c|} 
\hline
Class         & \multicolumn{1}{l|}{Train sample size} & \multicolumn{1}{l|}{Test sample size} \\ \hline
Bicycle       & 220                                    & 108                                   \\ \hline
Motorcycle    & 170                                    & 64                                    \\ \hline
Large vehicle & 250                                    & 131                                   \\ \hline
Small vehicle & 260                                    & 118                                   \\ \hline
Pedestrian    & 180                                    & 58                                    \\ \hline
\end{tabular}
\end{table}

\textbf{MNIST}: For the MNIST dataset, to simulate different views for each sensor, we rotate each image randomly between $0-180$ degrees.

\subsection{OTA multi-sensor wireless inference with digital sensors}
In this section, we will first evaluate the performance of over-the-air sensor fusion with $L_p$-norm inspired sensor fusion, using both fixed and trainable $p$, with exact digital computations at the edge sensors (e.g., without in-memory computing).

\begin{table*}[]
\centering
\caption{Network architecture for the multi-modal system with the ModelNet dataset and exact digital computations.} \label{tab_CNN3}
\vspace{0.15cm}
\begin{tabular}{|ccc|}
\hline
\multicolumn{3}{|c|}{Image: $224 \times 224 \times 3$}                                                                                                                                                                                                                                                                                                               \\ \hline
\multicolumn{1}{|c|}{}                                                                                     & \multicolumn{1}{c|}{\cellcolor[HTML]{C0C0C0}\textbf{Sensor 1}}                                                                       & \cellcolor[HTML]{C0C0C0}\textbf{Sensor 2}                                                                        \\ \cline{2-3} 
\multicolumn{1}{|c|}{}                                                                                     & \multicolumn{1}{c|}{\begin{tabular}[c]{@{}c@{}}$7 \times 7$ conv layer, \\ 32 channels, ReLU, \\ stride: 4, padding: 2\end{tabular}} & \begin{tabular}[c]{@{}c@{}}$5 \times 5$ conv layer, \\ 32 channels, ReLU, \\ stride: 4, padding: 2\end{tabular}  \\ \cline{2-3} 
\multicolumn{1}{|c|}{}                                                                                     & \multicolumn{1}{c|}{\begin{tabular}[c]{@{}c@{}}Max pooling with \\ kernel size: 3, stride: 2\end{tabular}}                           & \begin{tabular}[c]{@{}c@{}}Max pooling with \\ kernel size: 3, stride: 2\end{tabular}                            \\ \cline{2-3} 
\multicolumn{1}{|c|}{}                                                                                     & \multicolumn{1}{c|}{\begin{tabular}[c]{@{}c@{}}$5 \times 5$ conv layer, \\ 32 channels, ReLU, \\ stride: 1, padding: 2\end{tabular}} & \begin{tabular}[c]{@{}c@{}}$3 \times 3$ conv layer, \\ 16 channels, ReLU, \\ stride: 1, padding: 2\end{tabular}  \\ \cline{2-3} 
\multicolumn{1}{|c|}{}                                                                                     & \multicolumn{1}{c|}{\begin{tabular}[c]{@{}c@{}}Max pooling with \\ kernel size: 3, stride: 2\end{tabular}}                           & \begin{tabular}[c]{@{}c@{}}Max pooling with \\ kernel size: 3, stride: 2\end{tabular}                            \\ \cline{2-3} 
\multicolumn{1}{|c|}{}                                                                                     & \multicolumn{1}{c|}{Dropout layer ($p = 0.5$)}                                                                                       & Dropout layer ($p = 0.5$)                                                                                        \\ \cline{2-3} 
\multicolumn{1}{|c|}{}                                                                                     & \multicolumn{1}{c|}{\begin{tabular}[c]{@{}c@{}}Fully connected \\ layer (5408, 2048)\end{tabular}}                                   & \begin{tabular}[c]{@{}c@{}}Fully connected \\ layer (3136, 2048)\end{tabular}                                    \\ \cline{2-3} 
\multicolumn{1}{|c|}{}                                                                                     & \multicolumn{1}{c|}{\cellcolor[HTML]{C0C0C0}\textbf{Sensor 3}}                                                                       & \cellcolor[HTML]{C0C0C0}\textbf{Sensor 4}                                                                        \\ \cline{2-3} 
\multicolumn{1}{|c|}{}                                                                                     & \multicolumn{1}{c|}{\begin{tabular}[c]{@{}c@{}}$5 \times 5$ conv layer, \\ 16 channels, ReLU, \\ stride: 4, padding: 2\end{tabular}} & \begin{tabular}[c]{@{}c@{}}$3 \times 3$ conv layer, \\  16 channels, ReLU, \\ stride: 4, padding: 2\end{tabular} \\ \cline{2-3} 
\multicolumn{1}{|c|}{}                                                                                     & \multicolumn{1}{c|}{\begin{tabular}[c]{@{}c@{}}Max pooling with \\ kernel size: 3, stride: 2\end{tabular}}                           & \begin{tabular}[c]{@{}c@{}}Max pooling with \\ kernel size: 3, stride: 2\end{tabular}                            \\ \cline{2-3} 
\multicolumn{1}{|c|}{}                                                                                     & \multicolumn{1}{c|}{Dropout layer ($p = 0.5$)}                                                                                       & Dropout layer ($p = 0.5$)                                                                                        \\ \cline{2-3} 
\multicolumn{1}{|c|}{\multirow{-23}{*}{\begin{tabular}[c]{@{}c@{}}Front-end\\ (Sensor side)\end{tabular}}} & \multicolumn{1}{c|}{\begin{tabular}[c]{@{}c@{}}Fully connected \\ layer (11664,  2048)\end{tabular}}                                 & \begin{tabular}[c]{@{}c@{}}Fully connected \\ layer (12544,  2048)\end{tabular}                                  \\ \hline
\multicolumn{1}{|c|}{}                                                                                     & \multicolumn{2}{c|}{Fully connected layer (2048, 1024), ReLU activation}                                                                                                                                                                                \\ \cline{2-3} 
\multicolumn{1}{|c|}{\multirow{-2}{*}{\begin{tabular}[c]{@{}c@{}}Back-end\\ (Device side)\end{tabular}}}   & \multicolumn{2}{c|}{Fully connected layer (1024, 40), ReLU activation}                                                                                                                                                                                  \\ \hline
\end{tabular}
\end{table*}

\subsubsection{Approximating maximum operation for sensor fusion with the ModelNet dataset}

\begin{figure}[]
\centering
\includegraphics[width=0.9\linewidth]{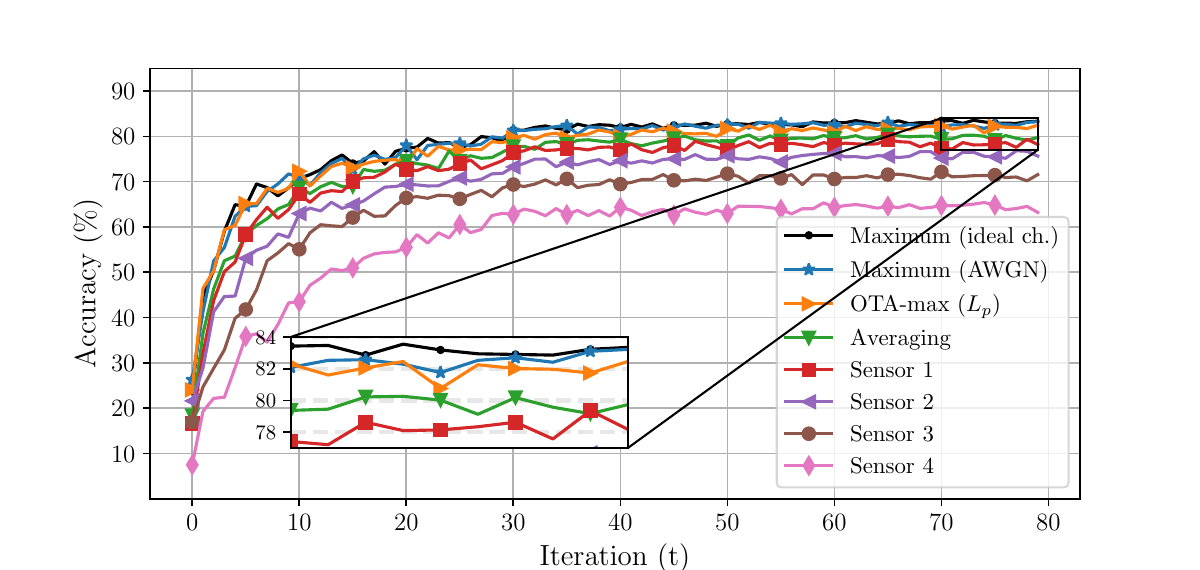}
         \caption{Inference accuracy for the ModelNet dataset with a multi-modal system with $M=4$ digital sensors each having  different computational capabilities.}
         \label{fig_CNN3}
\end{figure}

For multi-modal wireless inference with digital computations at the sensors using an $L_p$-norm inspired maximum approximation to obtain transformation-invariant features, we investigate the network structure given in Table \ref{tab_CNN3}.
We employ a mini-batch size of 10 and the Adam optimizer \cite{kingma2014adam} with an initial learning rate of $\eta = 10^{-4}$ over $T = 80$ iterations. 
{\color{black}{Note that in this section, we performed training over 80 iterations, during which the accuracy curves reach convergence, and learning stabilizes at a certain accuracy before reaching 80 iterations. However, the total number of iterations required to achieve convergence may vary depending on factors such as the dataset, its distribution, network structure, number of sensors, wireless channel properties, and other variables. As an automated approach, one could use a validation set to monitor accuracy after each iteration; when improvement ceases, early stopping can be applied to halt training.}} 
We consider the AWGN MAC channel with a 10 dB signal-to-noise ratio (SNR) and utilize the over-the-air approach described in Section \ref{ssec_Lp} with $p = 2$ for approximating the maximum operation.
The network consists of four sensors in which each sensor uses data randomly sampled from 12 views of the ModelNet dataset. In this simulation setup, the complexity of branches is adjusted according to the sensor capabilities, i.e.,
 we assume that the first sensor is the most powerful one in terms of computational capabilities while sensor four has the lowest capability.
 As shown in Fig. \ref{fig_CNN3}, the proposed over-the-air approximation achieves a performance close to that of the exact maximum, which can be considered as a performance upper bound for the given system model, while significantly outperforming the sensor fusion with averaging. The accuracy of single-sensor setups, which can be considered as baseline schemes, varies due to the multi-modality in the front-end network architectures and depends on the computational capabilities of the corresponding sensor, i.e., the complexity of the network branch. However, it is clear that sensor fusion helps improve the performance of all the sensors, with the accuracy of the one with the lowest capacity increasing the most, while still offering significant improvement for the most powerful one.

\subsubsection{Learnable sensor fusion with CARLA dataset}

\begin{table}[h]
\centering
\caption{Network architecture for the learnable sensor fusion for multi-sensor wireless inference.} \label{tab_learnable}
\vspace{0.15cm}
\begin{tabular}{|c|c|}
\hline
\multirow{4}{*}{\begin{tabular}[c]{@{}c@{}}Front-end\\ (Sensor side)\end{tabular}} & Image: $224 \times 224 \times 3$                                                                  \\ \cline{2-2} 
                                                                                   & \begin{tabular}[c]{@{}c@{}}
                                                                                   $3 \times 3$ convolutional layer, 32 channels, \\
                                                                                   ReLU, stride:4, padding: 2
                                                                                  \end{tabular} \\ \cline{2-2} 
                                                                                   & 2D maxpooling with kernelsize=3, stride=2                                                               \\ \cline{2-2} 
                                                                                   & dropout with probability 0.5                                                                                      \\ \hline
\multirow{4}{*}{\begin{tabular}[c]{@{}c@{}}Back-end\\ (Device-size)\end{tabular}}  & \begin{tabular}[c]{@{}c@{}} fully connected layer (23328, 8196)\\ ReLU\end{tabular}                                \\ \cline{2-2} 
                                                                                   & dropout with probability 0.5                                                                                        \\ \cline{2-2} 
                                                                                   & \begin{tabular}[c]{@{}c@{}}fully connected layer (8196, 2048)\\ ReLU\end{tabular}                                 \\ \cline{2-2} 
                                                                                   & Output layer: fully connected layer (2048, 40)                                                                    \\ \hline
\end{tabular}
\end{table}

For the learnable sensor fusion, we adopt the network structure outlined in Table \ref{tab_learnable} and employ $M = 5$ sensors for the custom-made CARLA dataset. During offline training, we utilize a mini-batch size of 10. The training process utilizes the Adam optimizer \cite{kingma2014adam} with an initial learning rate of $10^{-4}$ and runs for 80 iterations. For this dataset, we conduct training for 10 models and report the average test accuracy along with the associated one-standard-deviation interval. It is worth noting that both offline training and real-time inference involve a wireless channel connecting the sensors to the central processing device, modeled as an AWGN MAC. We initialize the learnable parameter $p$ as $0.95 + U[0, 0.1]$, where $U[0, 0.1]$ represents a uniform distribution between 0 and 0.1. This proposed approach is compared with three baselines: 1) sensor fusion by taking the exact maximum of all sensor's transmitted features, 2) feature averaging, 3) using only one sensor without sensor fusion.
Taking the exact maximum requires recovering all transmitted features from different sensors at the receiver side, necessitating orthogonal transmission and making the process costly and undesirable. On the other hand, both the learnable sensor fusion with the $L_p$-norm inspired function and feature averaging can be performed in an over-the-air manner with concurrent transmission, making them transmission efficient.
Furthermore, it is worth noting that with the help of the trainable parameter $p$ in learnable sensor fusion, one can cover both ends of the spectrum and approximate both averaging and exact maximum by adjusting $p$ values. Hence, the sensor fusion can be optimized for the given sensor network and data structure.

\begin{figure}[h]
     \centering
      %
    \includegraphics[width=0.9\linewidth]{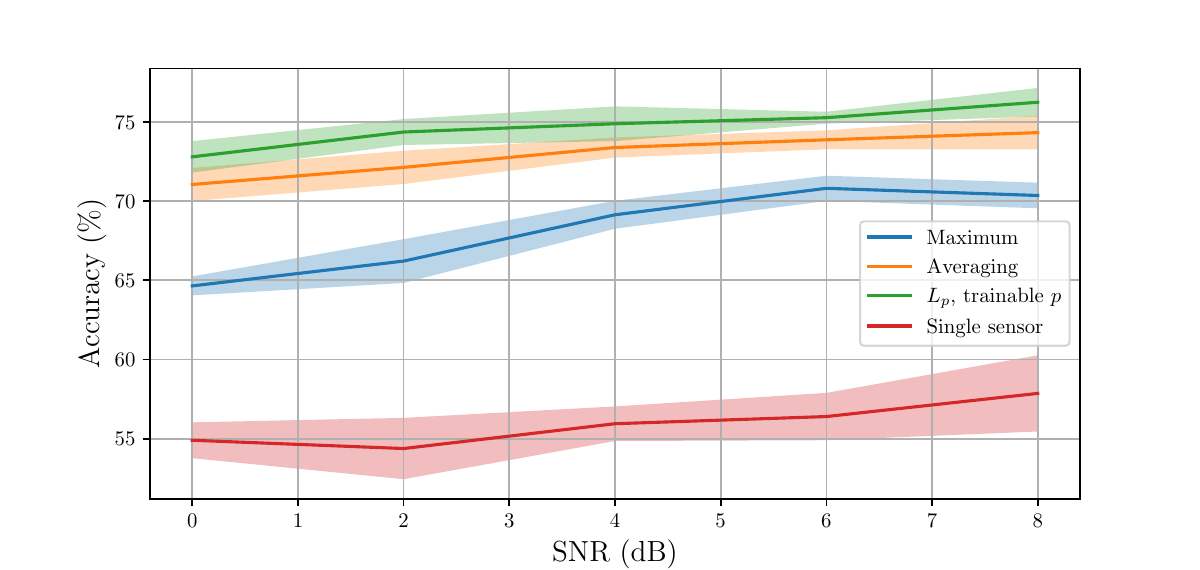}
        \caption{Inference accuracy for learnable sensor fusion using CARLA dataset with $M=5$ digital sensors with perfect (fixed) SNR training.}
         \label{CARLA_fixed}
\end{figure}

In Fig. \ref{CARLA_fixed}, we examine the perfectly known SNR scenario, where we assume that the channel SNR is available both during offline training and real-time processing. Consequently, we train and test the proposed approach and the baselines with exactly the same SNR settings. Specifically, we train multiple models with $\{0, 2, 4, 6, 8\}$ dB SNRs and test them with matching SNRs. As anticipated, using only one sensor without sensor fusion has the poorest performance, since it cannot exploit multiple data sources for the detected object. Moreover, the $L_p$-norm inspired sensor fusion achieves superior performance compared to both averaging and exact maximum across the entire SNR range. This observation highlights the benefits of introducing the learnable parameter $p$, which enables a sensor fusion method capable of generalizing better than averaging or exact maximum operations for the given setup. In this setup, it is noteworthy that the parameter $p$ converges to approximately $0.88$ {\color{black}{for 8 dB SNR}}, an average taken over $10$ models. Consequently, despite the final fusion method has different characteristics than averaging, it is closer to the averaging function (not to the maximum operation).

\begin{figure}[]
     \centering
      \includegraphics[width=0.9\linewidth]{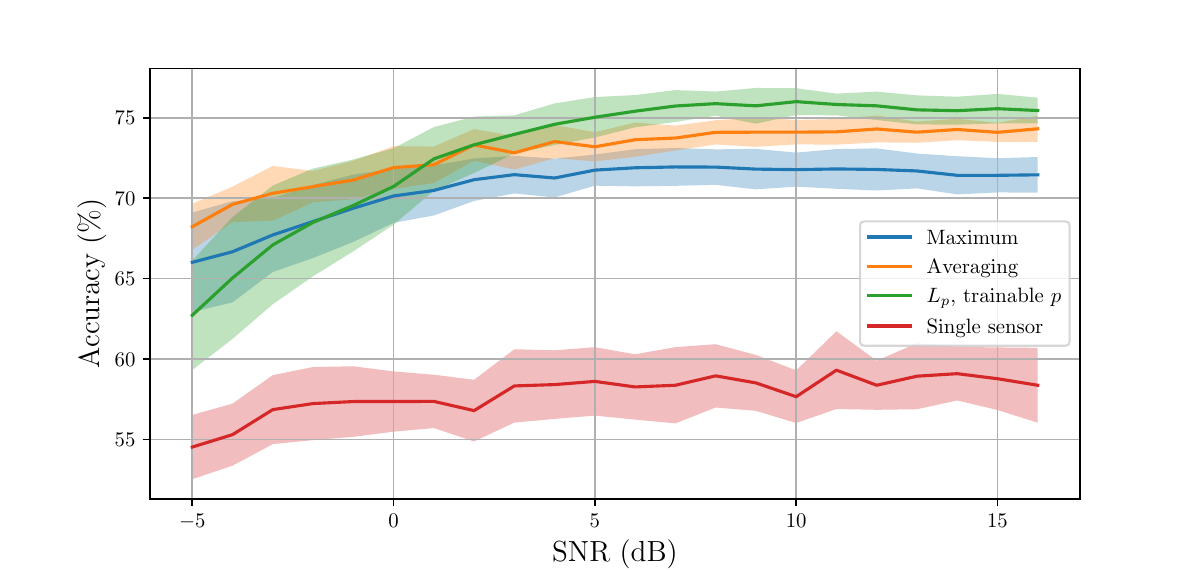}
        \caption{Inference accuracy for learnable sensor fusion using CARLA dataset with $M=5$ digital sensors with SNR-robust training.}
         \label{CARLA_robust}
\end{figure}

In Fig. \ref{CARLA_robust}, we investigate SNR-robust training for the same system. In this scenario, during both offline training and real-time inference, the only available information about the AWGN channel is the range of the SNR. The exact statistics of the channel is unknown. To address this, during each iteration of the training process, we sample an SNR value between $-5$ dB and $15$ dB (uniformly sampled in linear scale) and perform the training accordingly. Consequently, even though we lack perfect knowledge of the SNR, we can train a single network to accommodate the wide SNR range.
As depicted in Fig. \ref{CARLA_robust}, especially for moderate to high SNR values, our previous observation holds, confirming that the $L_p$-norm inspired sensor fusion with the trainable parameter remains a superior method compared to averaging and exact maximum. However, for low SNR values, e.g., for SNRs lower than 1 dB, averaging exhibits a slightly better performance compared to the proposed approach.
Furthermore, compared to the previous results, one does not need to train multiple models for each specific channel quality when employing the SNR-robust training approach. Instead, a single model is trained to adapt to the entire range of SNR values, making the system more versatile and efficient.

This observation indicates that the order of performance between averaging and exact maximum can vary based on the dataset, sensor distribution, or other system parameters. Nevertheless, with the learnable sensor fusion utilizing the $L_p$-norm inspired function, one can obtain a generalizable sensor fusion strategy, particularly beneficial for moderate to high SNR ranges. This demonstrates the adaptability and effectiveness of the proposed approach, making it suitable for diverse scenarios and system configurations.

We further note that the performance ranking of averaging and maximum differs between the custom-made CARLA dataset and the ModelNet dataset presented in the previous section. This observation suggests that the relative performance of averaging and exact maximum can vary depending on the dataset, sensor distribution, or other system parameters. These differences arise because the practical implementation of transformation-invariant features faces challenges, such as the need to position sensors throughout the entire scene and identify a set of transformations that meet the necessary properties. These limitations impose constraints on wireless inference systems, and even with transformation-invariant features, achieving the highest accuracy is not guaranteed. However, with learnable sensor fusion using the $L_p$-norm inspired function, a generalizable sensor fusion strategy can be achieved, which is particularly advantageous in moderate to high SNR ranges. This underscores the adaptability and effectiveness of the proposed approach, making it suitable for diverse scenarios and system configurations.

\subsection{Multi-sensor wireless inference with in-memory computing}
In this section, we provide some numerical results for multi-sensor wireless inference with in-memory computing and learnable sensor fusion using two different classification tasks with MNIST and ModelNet datasets. {\color{black}{Note that we also provide results with FP32 precision, which will serve as a performance benchmark and can be used as a comparative standard for assessing the performance of the memristor-based approach}}.

\subsubsection{MNIST}

\begin{table}
\centering
\caption{Network architecture for MNIST classification.}
\vspace{0.15cm}
\begin{tblr}{
  cells = {c},
  cell{1}{1} = {r=2}{},
  cell{3}{1} = {r=2}{},
  vlines,
  hline{1,3,5} = {-}{},
  hline{2,4}   = {2}{},
}
{Front-end\\(sensors)} & Image: $28 \times 28$   \\
                      & {5 $\times$ 5 convolutional layer, 10 channels,  ReLu, stride: $(1, 1)$} \\
{Back-end\\(device)}   & {Fully connected layer $(5760,50)$, ReLu} \\
                      & Output layer: fully connected $(50,10)$                       
\end{tblr}
\label{tab:MNIST_Architecture}
\end{table}

\begin{figure}
    \centering
    \includegraphics[width=0.8\linewidth]{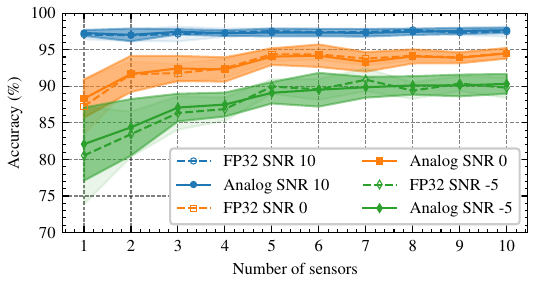}
    \caption{Test accuracy for MNIST classification with digital and analog sensors against number of sensors.}
    \label{fig:MNIST_Accuracay}
\end{figure}

In our initial investigation, we experiment with the MNIST classification, utilizing the network architecture given in Table \ref{tab:MNIST_Architecture}. We train the network under AWGN noise conditions, with SNR values set at $\{-5, 0, 10\}$ dB and sensors ranging from $M=1$ to $M=10$. To simulate different views for each sensor, we rotate each image randomly between $0-180$ degrees. Each configuration is trained with Adam optimizer \cite{kingma2014adam} at a learning rate of $\eta = 0.001$ for 200 epochs. The best model is saved for subsequent analog inference. Fig. \ref{fig:MNIST_Accuracay} shows the test accuracy for each scenario. Each data point represents the mean of 25 inferences, each derived from five programming trials, further analyzed over the communication channel five times. 

The results show that the analog sensor scenario exhibits performance that closely matches the digital implementation (FP32), thus supporting the advantages of integrating analog devices into neural networks with minimal performance degradation. The variability in the results, measured by the standard deviation, is attributable to two primary factors: first, the programming and read noise inherent in the memristive devices and second, the fluctuations in the wireless channel. An increase in the number of sensors and an improvement in the channel SNR contribute to more predictable performance with a significant reduction in standard deviation. Considering the extremely low energy usage of the memristive crossbars compared to traditional CPUs and GPUs, we argue that the proposed setup reduces the energy consumption in the sensors by a factor of at least $10^3$ \cite{kosters_benchmarking_2023}, encouraging the use of DNNs in edge devices for practical applications.

{\color{black}{

We note that in our simulations, there are several sources of randomness: channel noise, different noise types introduced by the memristors, inherent variability in neural network processing, and randomness in the input data (e.g., the rotation angle). As a result, while an increase in the number of sensors is generally expected to improve performance, small fluctuations are observed due to these random factors. Furthermore, the dataset used in this simulation is the MNIST dataset, with different random rotations applied for each sensor. Consequently, after reaching a certain number of sensors, all the data needed for accurate inference is effectively captured, making additional data redundant. Therefore, adding more sensors—for example, more than six in this specific case—does not lead to further performance improvement. With more complex datasets, however, a larger number of sensors may be necessary to achieve this saturation point.

}}

{\color{black}{

\textit{Energy Consumption:} For the MNIST classification task, we analyze the energy consumption of memristor-based sensors and compare it with Raspberry Pi-based sensors. Each memristor-based sensor processes only the convolutional layer, consisting of $260$ parameters. Using the energy consumption model introduced in \eqref{eq:upper_bound_sensors}, the energy consumption for the $m$-th sensor for one inference is:
\[
E_\text{m,max} = \text{(number of parameters per sensor)} \times g_\text{max} \times v_\text{max}^2 \times t.
\]
Substituting $260$ parameters, $g_\text{max} = 50 \, \mu\text{S}$, $v_\text{max} = 0.5 \, \text{V}$, and $t = 1 \, \text{ms}$, we have
\[
E_\text{m,max} = 260 \times 50 \times 10^{-6} \, \text{S} \times (0.5 \, \text{V})^2 \times 10^{-3} \, \text{s} = 3.25 \, \mu\text{J}.
\]
For $M = 10$ sensors operating in parallel:
\[
E_\text{total, max} = 10 \times 3.25 \, \mu\text{J} = 32.5 \, \mu\text{J}.
\]
In contrast, as outlined in Section \ref{sec:energy_consumption} with \eqref{eq:energy_rpi}, the energy consumption for Raspberry Pi-based sensors is:
\[
E_\text{RPi\_total} = M \times P_\text{RPi} \times t = 10 \times 15 \, \text{W} \times 10^{-3} \, \text{s} = 150 \, \text{mJ}.
\]
The memristor-based system achieves an energy efficiency improvement of approximately $4.6 \times 10^6$, demonstrating its exceptional suitability for energy-constrained edge applications. We note that our energy computation for memristor-based sensor systems accounts only for the matrix-vector multiplication operation. Additional operations may contribute to an increase in the total energy consumption of these systems.

}}

\subsubsection{ModelNet}

\begin{figure}
    \centering
    \includegraphics[width=0.8\linewidth]{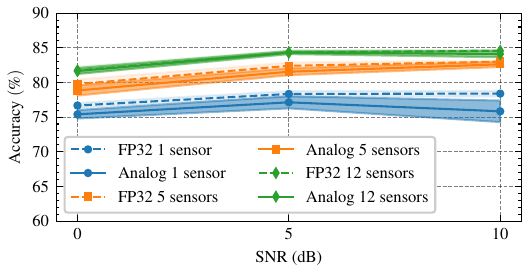}
    \caption{Test accuracy for ModelNet classification utilizing digital and analog sensors, with identical SNR during training.}
    \label{fig:ModelNet_Accuracy_same_SNR}
\end{figure}

For our subsequent analysis, we selected the ModelNet \cite{mvcnn} dataset, with the network architecture given in Table \ref{tab_learnable}. %
The network performance is evaluated for three different scenarios of training and test channels 1) Perfect channel-state-information (CSI), 2) CSI mismatch and, 3) Robust training. Each configuration is trained with Adam optimizer, with a learning rate set at $\eta = 0.0001$, for 100 epochs, incorporating configurations of 1, 5, and 12 sensors. Each model is tested for channels with SNRs $\{0, 5, 10\}$ dB.

\begin{figure}
    \centering
    \includegraphics[width=0.8\linewidth]{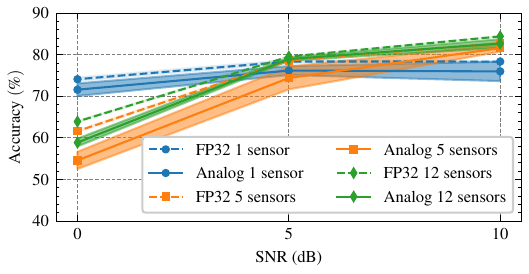}
    \caption{Test accuracy for ModelNet classification utilizing digital and analog sensors, with 10dB SNR during training.}
    \label{fig:ModelNet_10db_channel}
\end{figure}

\textit{Perfect CSI:} Fig. \ref{fig:ModelNet_Accuracy_same_SNR} presents the test accuracy for analog and digital sensors under conditions where the SNR for both testing and training phases is identical. As expected, increasing the number of sensors improves the performance for all channel conditions. Moreover, an observed decrement of 3-5\% in accuracy is seen when employing analog sensors. This phenomenon is attributed to the increased complexity of the network, which in turn introduces a greater chance for errors within the network weights as each weight in the sensor layers is subjected to a weight-dependent noise as defined in \eqref{eqn:n_prog}.

\begin{figure}[]
    \centering
    \includegraphics[width=0.8\linewidth]{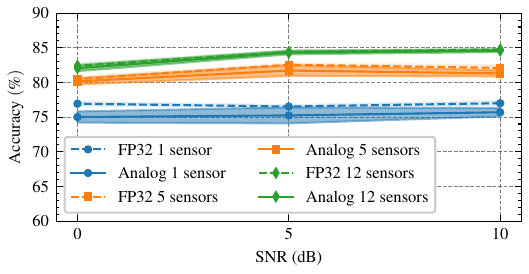}
    \caption{Test accuracy for ModelNet classification utilizing digital and analog sensors, with a random SNR during training.}
    \label{fig:ModelNet_random_0-10dB_channel}
\end{figure}

\textit{CSI mismatch:} Fig. \ref{fig:ModelNet_10db_channel} display the test accuracy results under various SNR conditions where the models are tested at an SNR different than the training conditions. The results clearly indicate that the best performance is achieved when the training and test channel conditions are matched, while a significant loss in performance is observed when there is a mismatch. Specifically, the model trained at 10 dB SNR and 12 sensors achieved an accuracy of 84\% when tested at 10 dB, but the accuracy dropped to 78\% and 64\% when tested at 5 dB and 0 dB, respectively. This strengthens our findings from Fig. \ref{fig:ModelNet_Accuracy_same_SNR} that incorporating perfect CSI information during the training phase enhances the performance significantly. However, achieving good performance across a range of channel conditions would require training a model for each possible channel condition, which is not feasible for practical reasons.

\textit{Robust training:} To avoid training multiple models for different channel conditions, we propose a training approach where each epoch uses a uniformly selected SNR between 0 dB and 10 dB. In Fig. \ref{fig:ModelNet_random_0-10dB_channel}, we show the test performance under various channel SNR values. Each data point represents the mean of 25 inferences from five distinct programming trials, each inferred five times. Training with random SNR per epoch increases adaptability to fluctuating channel conditions enhancing robustness and ensuring reliable performance even with unpredictable channel conditions. By incorporating this randomized training, we avoid the impracticality of training separate models for each channel condition, offering a feasible solution for real-world applications.

\begin{figure}[]
    \centering
    \includegraphics[width=0.8\linewidth]{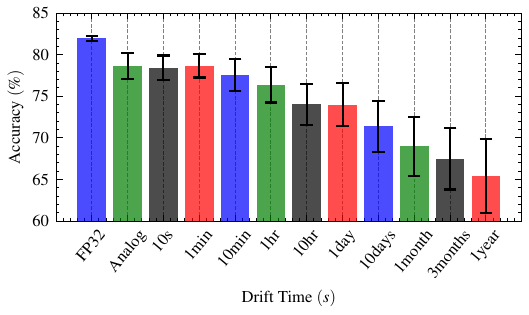}
    \caption{Accuracy vs drift time for a network with 5 sensors at 10 dB SNR.}
\label{fig:accuracy_vs_drift_time_5_sensors_10.0_dB_SNR}
\end{figure}

Fig. \ref{fig:accuracy_vs_drift_time_5_sensors_10.0_dB_SNR} illustrates the test accuracy for various time drift values. The model is trained with five sensors with an SNR range of 0 dB to 10 dB, and then tested at 10 dB SNR. Each bar in Fig. \ref{fig:accuracy_vs_drift_time_5_sensors_10.0_dB_SNR} represents the mean and standard deviation over 1000 runs, with each run comprising 100 drift trials, each inferred over the channel 10 times. This setup simulates long-term weight drift in PCM type devices within memristive neural networks as described in \eqref{eqn:g_drift}. Without conductance drift, accuracy decreases by only 4-5\%. However, with conductance drift, the accuracy gradually declines and reaches just above 65\% after one year. This decline reflects the real-world scenario and underscores the need for continuous monitoring and reprogramming to maintain performance. This study helps identify the optimal reprogramming interval for a sustainable classification accuracy. With these results, we verify that even for long-term usage of a memristive neural network, there is only minimal performance loss, while the analog computations provide significant energy savings. Hence, considering the trade-off between energy efficiency and performance, memristive neural networks offer a feasible solution for edge devices.

{\color{black}{
Note that we observe some variations in the numerical results. While memristor noise contributes to this variability, it can be mitigated through approaches such as noise-aware training, as proposed in \cite{kariyappa2021noise}. However, the variations in our results are not solely due to memristor noise; they are also influenced by factors like channel noise (SNR value) and the number of sensors. This variability can be significantly reduced by using higher SNR values and increasing the number of sensors. Moreover, inherent randomness in stochastic machine learning approaches adds to the variations, making it impossible to eliminate them entirely. In future work, we plan to explore techniques such as error correction and noise compensation algorithms to further enhance computational robustness against memristor noise. }}

We conclude this section by noting that in our simulations, we consider an adaptable over-the-air combining function inspired by the $L_p$-norm, where the parameter $p$ is treated as a trainable network parameter during training. With a fixed $p$, one needs to decide on the sensor fusion method manually, e.g., average or maximum, before training. In contrast, the proposed approach allows dynamic adjustment and optimization of the sensor fusion method without manual parameter assignment, suiting diverse scenarios.

\section{Conclusions} \label{sec:conc}

We have investigated a novel multi-sensor fusion framework utilizing analog memristor neural networks. The implementation of transformation-invariant $L_p$-norm inspired sensor fusion significantly reduces communication costs with OTA sensor fusion. We further optimize the system by introducing a learnable parameter, which enhances flexibility. This parameter allows deep neural networks to dynamically adjust their sensor fusion method, ranging from averaging to maximum, to suit diverse scenarios and system configurations. Additionally, in-memory computing greatly improves energy efficiency compared to traditional CPU and GPU computations, with only minimal performance sacrifice. This offers an efficient solution for complex tasks in distributed environments.

\section{Acknowledgment}
The authors would like to acknowledge the efforts of Mert Kalfa, Yusuf Kesmen, and Yiğit Küçük for their invaluable contributions in the development of the custom-made CARLA dataset.

\bibliographystyle{elsarticle-num}
\bibliography{bibs_edge}

\begin{thebibliography}{10}
\expandafter\ifx\csname url\endcsname\relax
  \def\url#1{\texttt{#1}}\fi
\expandafter\ifx\csname urlprefix\endcsname\relax\def\urlprefix{URL }\fi
\expandafter\ifx\csname href\endcsname\relax
  \def\href#1#2{#2} \def\path#1{#1}\fi

\bibitem{lecun1989optimal}
Y.~LeCun, J.~Denker, S.~Solla, Optimal brain damage, Advances in Neural
  Information Processing Systems 2 (1990) 1--8.

\bibitem{hassibi1993optimal}
B.~Hassibi, D.~Stork, G.~Wolff, Optimal brain surgeon: Extensions and
  performance comparisons, Advances in Neural Information Processing Systems 6
  (1993) 1--8.

\bibitem{blalock2020state}
D.~Blalock, J.~J. Gonzalez~Ortiz, J.~Frankle, J.~Guttag, What is the state of
  neural network pruning?, Proceedings of Machine Learning and Systems 2 (2020)
  129--146.

\bibitem{shao_bottlenet_2020}
J.~Shao, J.~Zhang, {BottleNet}++: {An} {End}-to-{End} {Approach} for {Feature}
  {Compression} in {Device}-{Edge} {Co}-{Inference} {Systems}, in: {IEEE}
  {Int.} {Conf.} on {Commn.} {Workshops}, Dublin, Ireland, 2020, pp. 1--6.
\newblock \href {https://doi.org/10.1109/ICCWorkshops49005.2020.9145068}
  {\path{doi:10.1109/ICCWorkshops49005.2020.9145068}}.

\bibitem{Jankowski}
M.~Jankowski, D.~Gündüz, K.~Mikolajczyk, Joint device-edge inference over
  wireless links with pruning, in: IEEE 21st International Workshop on Signal
  Processing Advances in Wireless Communications (SPAWC), Atlanta, GA, USA,
  2020, pp. 1--5.
\newblock \href {https://doi.org/10.1109/SPAWC48557.2020.9154306}
  {\path{doi:10.1109/SPAWC48557.2020.9154306}}.

\bibitem{image_retrieval}
M.~Jankowski, D.~Gündüz, K.~Mikolajczyk, Wireless image retrieval at the
  edge, IEEE J. Sel. Areas Commun. 39~(1) (2021) 89--100.
\newblock \href {https://doi.org/10.1109/JSAC.2020.3036955}
  {\path{doi:10.1109/JSAC.2020.3036955}}.

\bibitem{Branchy-GNN}
J.~Shao, H.~Zhang, Y.~Mao, J.~Zhang, Branchy-{GNN}: A device-edge co-inference
  framework for efficient point cloud processing, in: IEEE International
  Conference on Acoustics, Speech and Signal Processing (ICASSP), Toronto, ON,
  Canada, 2021, pp. 8488--8492.
\newblock \href {https://doi.org/10.1109/ICASSP39728.2021.9414831}
  {\path{doi:10.1109/ICASSP39728.2021.9414831}}.

\bibitem{9606569}
M.~Lee, G.~Yu, H.~Dai, Decentralized inference with graph neural networks in
  wireless communication systems, IEEE Transactions on Mobile Computing 22~(5)
  (2023) 2582--2598.
\newblock \href {https://doi.org/10.1109/TMC.2021.3125793}
  {\path{doi:10.1109/TMC.2021.3125793}}.

\bibitem{lee2022privacy}
M.~Lee, G.~Yu, H.~Dai, Privacy-preserving decentralized inference with graph
  neural networks in wireless networks, IEEE Transactions on Wireless
  Communications 23~(1) (2024) 543--558.
\newblock \href {https://doi.org/10.1109/TWC.2023.3279442}
  {\path{doi:10.1109/TWC.2023.3279442}}.

\bibitem{shao_task}
J.~Shao, Y.~Mao, J.~Zhang, Learning task-oriented communication for edge
  inference: An information bottleneck approach, IEEE J. Sel. Areas Commun.
  40~(1) (2022) 197--211.
\newblock \href {https://doi.org/10.1109/JSAC.2021.3126087}
  {\path{doi:10.1109/JSAC.2021.3126087}}.

\bibitem{shao_multi}
J.~Shao, Y.~Mao, J.~Zhang, Task-oriented communication for multidevice
  cooperative edge inference, IEEE Trans. on Wireless Commun. 22~(1) (2023)
  73--87.
\newblock \href {https://doi.org/10.1109/TWC.2022.3191118}
  {\path{doi:10.1109/TWC.2022.3191118}}.

\bibitem{tishby2000information}
N.~Tishby, F.~C. Pereira, W.~Bialek, The information bottleneck method, arXiv
  preprint physics/0004057 (2000).

\bibitem{IB_deep}
N.~Tishby, N.~Zaslavsky, Deep learning and the information bottleneck
  principle, in: IEEE Information Theory Workshop (ITW), Jerusalem, Israel,
  2015, pp. 1--5.
\newblock \href {https://doi.org/10.1109/ITW.2015.7133169}
  {\path{doi:10.1109/ITW.2015.7133169}}.

\bibitem{9834591}
S.~F. Yilmaz, B.~Hasırcıoğlu, D.~Gündüz, Over-the-air ensemble inference
  with model privacy, in: IEEE International Symposium on Information Theory
  (ISIT), Espoo, Finland, 2022, pp. 1265--1270.

\bibitem{li2018edge}
E.~Li, Z.~Zhou, X.~Chen, Edge intelligence: On-demand deep learning model
  co-inference with device-edge synergy, in: Proceedings of the 2018 Workshop
  on Mobile Edge Communications, Budapest Hungary, 2018, pp. 31--36.
\newblock \href {https://doi.org/10.1145/3229556.3229562}
  {\path{doi:10.1145/3229556.3229562}}.

\bibitem{liu2019edge}
L.~Liu, H.~Li, M.~Gruteser, Edge assisted real-time object detection for mobile
  augmented reality, in: The 25th Annual Int. Conf. on Mobile Computing and
  Networking, Los Cabos, Mexico, 2019, pp. 1--16.
\newblock \href {https://doi.org/10.1145/3300061.3300116}
  {\path{doi:10.1145/3300061.3300116}}.

\bibitem{kang2017neurosurgeon}
Y.~Kang, J.~Hauswald, C.~Gao, A.~Rovinski, T.~Mudge, J.~Mars, L.~Tang,
  Neurosurgeon: Collaborative intelligence between the cloud and mobile edge,
  ACM SIGARCH Computer Architecture News 45~(1) (2017) 615--629.
\newblock \href {https://doi.org/10.1145/3093337.3037698}
  {\path{doi:10.1145/3093337.3037698}}.

\bibitem{li2018jalad}
H.~Li, C.~Hu, J.~Jiang, Z.~Wang, Y.~Wen, W.~Zhu, {JALAD}: Joint accuracy-and
  latency-aware deep structure decoupling for edge-cloud execution, in: IEEE
  24th International Conference on Parallel and Distributed Systems (ICPADS),
  Singapore, 2018, pp. 671--678.
\newblock \href {https://doi.org/10.1109/PADSW.2018.8645013}
  {\path{doi:10.1109/PADSW.2018.8645013}}.

\bibitem{8861554}
M.~Kenyeres, J.~Kenyeres, Applicability of generalized metropolis-hastings
  algorithm in wireless sensor networks, in: IEEE International Conference on
  Smart Technologies (EUROCON), Novi Sad, Serbia, 2019, pp. 1--5.

\bibitem{von_neumann_first_1993}
J.~von Neumann, First draft of a report on the {EDVAC}, IEEE Annals of the
  History of Computing 15~(4) (1993) 27--75.
\newblock \href {https://doi.org/10.1109/85.238389}
  {\path{doi:10.1109/85.238389}}.

\bibitem{chua_memristormissing_1971}
L.~O. Chua, Memristor—{The} {Missing} {Circuit} {Element}, IEEE Transactions
  on Circuit Theory 18~(5) (1971) 507--519.
\newblock \href {https://doi.org/10.1109/TCT.1971.1083337}
  {\path{doi:10.1109/TCT.1971.1083337}}.

\bibitem{strukov_missing_2008}
D.~B. Strukov, G.~S. Snider, D.~R. Stewart, R.~S. Williams, The missing
  memristor found, Nature 2008 453:7191 453~(7191) (2008) 80--83.
\newblock \href {https://doi.org/10.1038/nature06932}
  {\path{doi:10.1038/nature06932}}.

\bibitem{zhang_brain-inspired_2020}
Y.~Zhang, Z.~Wang, J.~Zhu, Y.~Yang, M.~Rao, W.~Song, Y.~Zhuo, X.~Zhang, M.~Cui,
  L.~Shen, R.~Huang, J.~J. Yang, Brain-inspired computing with memristors:
  {Challenges} in devices, circuits, and systems, Applied Physics Reviews 7~(1)
  (Mar. 2020).
\newblock \href {https://doi.org/10.1063/1.5124027}
  {\path{doi:10.1063/1.5124027}}.

\bibitem{kosters_benchmarking_2023}
D.~J. Kösters, B.~A. Kortman, I.~Boybat, E.~Ferro, S.~Dolas, R.~Ruiz~de
  Austri, J.~Kwisthout, H.~Hilgenkamp, T.~Rasing, H.~Riel, A.~Sebastian,
  S.~Caron, J.~H. Mentink, Benchmarking energy consumption and latency for
  neuromorphic computing in condensed matter and particle physics, APL Machine
  Learning 1~(1) (2023) 016101.
\newblock \href {https://doi.org/10.1063/5.0116699}
  {\path{doi:10.1063/5.0116699}}.

\bibitem{isik_neural_2023}
B.~Isik, K.~Choi, X.~Zheng, T.~Weissman, S.~Ermon, H.-S.~P. Wong, A.~Alaghi,
  Neural network compression for noisy storage devices, ACM Trans. Embed.
  Comput. Syst. 22~(3) (May 2023).
\newblock \href {https://doi.org/10.1145/3588436} {\path{doi:10.1145/3588436}}.

\bibitem{sebastian_memory_2020}
A.~Sebastian, M.~Le~Gallo, R.~Khaddam-Aljameh, E.~Eleftheriou, Memory devices
  and applications for in-memory computing, Nature Nanotechnology 15~(7) (2020)
  529--544.
\newblock \href {https://doi.org/10.1038/s41565-020-0655-z}
  {\path{doi:10.1038/s41565-020-0655-z}}.

\bibitem{rault2014energy}
T.~Rault, A.~Bouabdallah, Y.~Challal, Energy efficiency in wireless sensor
  networks: A top-down survey, Computer networks 67 (2014) 104--122.

\bibitem{6182561}
M.~Doudou, D.~Djenouri, N.~Badache, Survey on latency issues of asynchronous
  mac protocols in delay-sensitive wireless sensor networks, IEEE
  Communications Surveys \& Tutorials 15~(2) (2013) 528--550.

\bibitem{prauzek2018energy}
M.~Prauzek, J.~Konecny, M.~Borova, K.~Janosova, J.~Hlavica, P.~Musilek, Energy
  harvesting sources, storage devices and system topologies for environmental
  wireless sensor networks: A review, Sensors 18~(8) (2018) 2446.

\bibitem{khashan2021automated}
O.~A. Khashan, R.~Ahmad, N.~M. Khafajah, An automated lightweight encryption
  scheme for secure and energy-efficient communication in wireless sensor
  networks, Ad Hoc Networks 115 (2021) 102448.

\bibitem{tegin_transformation-invariant_2023}
B.~Tegin, T.~M. Duman, Transformation-{Invariant} {Over}-the-{Air} {Combining}
  for {Multi}-{Sensor} {Wireless} {Inference}, in: {IEEE} {Global}
  {Communications} {Conf.}, Kuala Lumpur, Malaysia, 2023, pp. 2937--2942.
\newblock \href {https://doi.org/10.1109/GLOBECOM54140.2023.10436926}
  {\path{doi:10.1109/GLOBECOM54140.2023.10436926}}.

\bibitem{teerapittayanon2017distributed}
S.~Teerapittayanon, B.~McDanel, H.-T. Kung, Distributed deep neural networks
  over the cloud, the edge and end devices, in: IEEE 37th international
  conference on distributed computing systems (ICDCS), Atlanta, GA, USA, 2017,
  pp. 328--339.
\newblock \href {https://doi.org/10.1109/ICDCS.2017.226}
  {\path{doi:10.1109/ICDCS.2017.226}}.

\bibitem{laptev2015transformation}
D.~Laptev, J.~M. Buhmann, Transformation-invariant convolutional jungles, in:
  Proceedings of the IEEE Conference on Computer Vision and Pattern
  Recognition, 2015, pp. 3043--3051.

\bibitem{laptev2016ti}
D.~Laptev, N.~Savinov, J.~M. Buhmann, M.~Pollefeys, {TI}-pooling:
  transformation-invariant pooling for feature learning in convolutional neural
  networks, in: Proceedings of the IEEE conference on computer vision and
  pattern recognition, 2016, pp. 289--297.

\bibitem{mvcnn}
H.~Su, S.~Maji, E.~Kalogerakis, E.~Learned-Miller, Multi-view convolutional
  neural networks for {3D} shape recognition, in: Proceedings of the IEEE
  International Conference on Computer Vision, Santiago, Chile, 2015, pp.
  945--953.

\bibitem{carbajal_memristor_2015}
J.~P. Carbajal, J.~Dambre, M.~Hermans, B.~Schrauwen, Memristor models for
  machine learning, Neural Computation 27~(3) (2015) 725--747.
\newblock \href {https://doi.org/{10.1162/NECO\_a\_00694}}
  {\path{doi:{10.1162/NECO\_a\_00694}}}.

\bibitem{nandakumar_phase-change_2018}
S.~R. Nandakumar, M.~Le~Gallo, I.~Boybat, B.~Rajendran, A.~Sebastian,
  E.~Eleftheriou, A phase-change memory model for neuromorphic computing,
  Journal of Applied Physics 124~(15) (2018) 152135.
\newblock \href {https://doi.org/10.1063/1.5042408}
  {\path{doi:10.1063/1.5042408}}.

\bibitem{noauthor_inference_nodate}
\href{https://aihwkit.readthedocs.io/en/latest/pcm_inference.html#references-pcm}{Inference
  and {PCM} {Statistical} {Model} — {IBM} {Analog} {Hardware} {Acceleration}
  {Kit} 0.9.0 documentation}.
\newline\urlprefix\url{https://aihwkit.readthedocs.io/en/latest/pcm_inference.html#references-pcm}

\bibitem{9535447}
G.~Zhu, J.~Xu, K.~Huang, S.~Cui, Over-the-air computing for wireless data
  aggregation in massive {IoT}, IEEE Wireless Communications 28~(4) (2021)
  57--65.
\newblock \href {https://doi.org/10.1109/MWC.011.2000467}
  {\path{doi:10.1109/MWC.011.2000467}}.

\bibitem{9382114}
M.~M. Amiri, T.~M. Duman, D.~Gündüz, S.~R. Kulkarni, H.~V. Poor, Blind
  federated edge learning, IEEE Transactions on Wireless Communications 20~(8)
  (2021) 5129--5143.
\newblock \href {https://doi.org/10.1109/TWC.2021.3065920}
  {\path{doi:10.1109/TWC.2021.3065920}}.

\bibitem{10018930}
B.~Tegin, T.~M. Duman, Federated learning with over-the-air aggregation over
  time-varying channels, IEEE Transactions on Wireless Communications 22~(8)
  (2023) 5671--5684.
\newblock \href {https://doi.org/10.1109/TWC.2023.3235894}
  {\path{doi:10.1109/TWC.2023.3235894}}.

\bibitem{8371243}
L.~Chen, N.~Zhao, Y.~Chen, F.~R. Yu, G.~Wei, Over-the-air computation for {IoT}
  networks: Computing multiple functions with antenna arrays, IEEE Internet of
  Things Journal 5~(6) (2018) 5296--5306.
\newblock \href {https://doi.org/10.1109/JIOT.2018.2843321}
  {\path{doi:10.1109/JIOT.2018.2843321}}.

\bibitem{van2001art}
D.~A. Van~Dyk, X.-L. Meng, The art of data augmentation, J. of Computational
  and Graphical Statistics 10~(1) (2001) 1--50.
\newblock \href {https://doi.org/10.1198/10618600152418584}
  {\path{doi:10.1198/10618600152418584}}.

\bibitem{le_gallo_collective_2018}
M.~Le~Gallo, D.~Krebs, F.~Zipoli, M.~Salinga, A.~Sebastian, Collective
  {Structural} {Relaxation} in {Phase}-{Change} {Memory} {Devices}, Advanced
  Electronic Materials 4~(9) (2018) 1700627.
\newblock \href {https://doi.org/10.1002/aelm.201700627}
  {\path{doi:10.1002/aelm.201700627}}.

\bibitem{fugazza_random_2010}
D.~Fugazza, D.~Ielmini, S.~Lavizzari, A.~L. Lacaita, Random telegraph signal
  noise in phase change memory devices, in: {IEEE} {Int.} {Reliability}
  {Physics} {Symp.}, Anaheim, CA, USA, 2010, pp. 743--749.
\newblock \href {https://doi.org/10.1109/IRPS.2010.5488741}
  {\path{doi:10.1109/IRPS.2010.5488741}}.

\bibitem{kariyappa2021noise}
S.~Kariyappa, H.~Tsai, K.~Spoon, S.~Ambrogio, P.~Narayanan, C.~Mackin, A.~Chen,
  M.~Qureshi, G.~W. Burr, {Noise-resilient DNN:} tolerating noise in
  {PCM-based} {AI} accelerators via noise-aware training, IEEE Transactions on
  Electron Devices 68~(9) (2021) 4356--4362.

\bibitem{ali2024learning}
M.~A. Ali, Learning and inference for wireless communications applications
  using in-memory analog computing, Ph.D. thesis, Bilkent University (2024).

\bibitem{Wu_2015_CVPR}
Z.~Wu, S.~Song, A.~Khosla, F.~Yu, L.~Zhang, X.~Tang, J.~Xiao, {3D} shapenets: A
  deep representation for volumetric shapes, in: Proceedings of the IEEE
  Conference on Computer Vision and Pattern Recognition (CVPR), 2015, pp.
  1912--1920.

\bibitem{dosovitskiy2017carla}
A.~Dosovitskiy, G.~Ros, F.~Codevilla, A.~Lopez, V.~Koltun, {CARLA}: An open
  urban driving simulator, in: Conference on robot learning, PMLR, 2017, pp.
  1--16.

\bibitem{kingma2014adam}
D.~P. Kingma, J.~Ba, Adam: A method for stochastic optimization, arXiv preprint
  arXiv:1412.6980 (2014).

\end{thebibliography}

\end{document}